# UPANets: Learning from the Universal Pixel Attention Networks

**Ching-Hsun Tseng[1], Shin-Jye Lee[2], Jia-Nan Feng[3], Shengzhong Mao[1], Yu-Ping Wu[1], Jia-Yu Shang[1], Mou-Chung Tseng[4], Xiao-Jun Zeng[1]**

[1]Department of Computer Science, The University of Manchester, Manchester, UK.
[2]Institute of Management of Technology, National Chiao Tung University, Hinschu, Taiwan.
[3]School of Software, Yunnan University, Kunming, China.
[4]College of Management, National Taipei University of Technology, Taipei, Taiwan

Corresponding author: Shin-Jye Lee (e-mail: camhero@gmail.com)

**ABSTRACT**

Among image classification, skip and densely-connection-based networks have dominated most leaderboards. Recently, from the successful development of multi-head attention in natural language processing, it is sure that now is a time of either using a Transformer-like model or hybrid CNNs with attention. However, the former need a tremendous resource to train, and the latter is in the perfect balance in this direction. In this work, to make CNNs handle global and local information, we proposed UPANets, which equips channel-wise attention with a hybrid skip-densely-connection structure. Also, the extreme-connection structure makes UPANets robust with a smoother loss landscape. In experiments, UPANets surpassed most well-known and widely-used SOTAs with an accuracy of 96.47% in Cifar-10, 80.29% in Cifar-100, and 67.67% in Tiny Imagenet. Most importantly, these performances have high parameters efficiency and only trained in one customer-based GPU. We share implementing code of UPANets in https://github.com/hanktseng131415go/UPANets.

## I. INTRODUCTION

The field of Computer vision has experienced a range of trends in a decade. Except for fundamental machine learning methods [1] and deep fully-connected convolutional neural networks [2], the introducing models of [3-5] [6] [7] in Imagnet competition has boomed the image classification. A variety of CNN-based model with residual, also known as skip-connection, networks [8-15] has conquered Cifar-10, Cifar-100, and Imagenet. Although some discussions and works, such as [17], mentioned convolutional layer could capture local characteristic and global profile if CNNs were in deep structure, the authors of [18] have argued the duty to capture global pattern is contributed with an attention mechanism. Also, because [18] has opened a path of applying pure multi-head attention from Transformer to image classification, some works, such as [19, 20], started to apply pure attention in computer vision. Not only toward computer vision, [21] utilized a sparse attention mechanism to make time-series forecasting more efficient. Therefore, the usage of attention does popularize in many categories nowadays. However, we have also noticed that most attention-based methods need powerful GPUs with large exclusive CUDA memory because generating the query, key, and value needs at least three times more resource than simply using one multi-layer perceptron. If we are facing computer vision with high resolution and many channels, the needed resource is unprecedented.

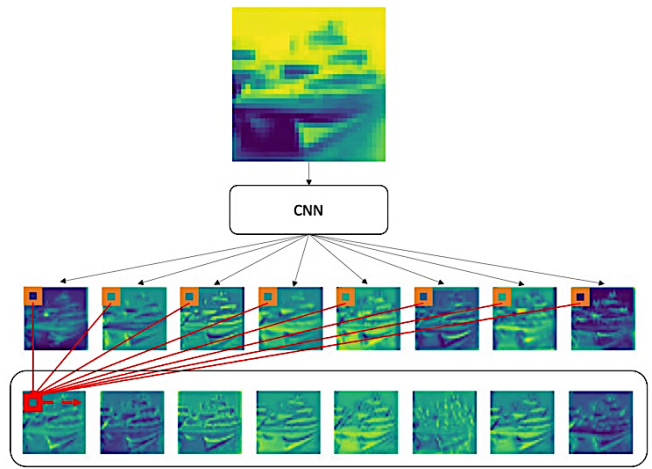

**Figure 1.** Channel pixel attention process and samples. The image on the top is an original sampled image from Cifar-10. The feature maps in the middle line are the outputs from the CNNs before CPA. On the bottom line are the samples from CPA. The red square is the weighted pixel sum from each orange square pixel in the same position.

In this regard, we want to endow the already excellent and efficient CNN-based networks to capture global information with learnable parameter and lesser resource than multi-head attention, so we proposed Channel-wise Pixel Attention CPA to make global pattern learning as **Figure 1**. Also, as residual neural networks have shined in image classification, densely-connection CNNs [22] also occupy the aforementioned well-known image datasets leaderboard. With the observation in [16], we improve performance by proposing another hybrid skip-densely-connection structure similar to dual-path networks [23]. By integrating proposed methods into a networks, our UPANets can additionally process universal pixels with CNNs and CPA, reuse feature maps by densely-connection, residual learning with skip-connection, and create a smooth learning landscape toward spatial pixel attention with extreme connection.

We first discuss an essential background and current trend toward image classification with merits and flaws in **I. INTRODUCTION** in this work. The contributions which have been brought by proposed methods are also listed in here. Then, in **II. RELTEDWORK,** the well-known and vital observation toward image classification and this work were mentioned with a critical analysis. Then a range of the proposed methods and the structure about UPANets were in **III. UPANets.** Moreover, comparing performance in terms of every proposed method in well-known datasets can be seen in **IV. EXPERIMENT** and **V. CONCLUSION.** Lastly, extra findings about UPANets experiments were in **Appendix.** The contributions from this work are:
- Channel pixel attention, which helps form complex features even in shallow depth with fewer parameters.
- Spatial pixel attention, which helps to learn spatial information.
- Hybrid skip-densely connection, which makes CNNs reuse feature with a deep structure.
- Extreme connection, which can generate a smooth loss landscape.
- A competitive image classification model surpassed well-known, also widely-used SOTAs in Cifar-10, Cifar-100, and Tiny Imagenet.

**II. RELTEDWORK**
Since the introducing of skip connection of ResNets [6], we have witnessed a surge in computer vision toward creating a smooth loss landscape. The skip connection has offered a great path to let deep learning fulfil the true meaning of dee. Most importantly, it prevents overfitting. The visualization of loss landscape [16] has proven one of the reasons that why simply applying skip connection can boost accuracy. Also, DenseNets [22] has shown another method to connect original and outputting information. [16] also has shown that using densely-connection makes the loss landscape smoother than ResNets. Following that, dual path networks [23] combining the merit of adding residual as ResNets and the inheriting input information as DenseNets. Not only that, the Deep layer aggregation model [24] similarly used dense connectivity to build a tree-based structure toward fusing images and image detection. Among the development of creating a smooth loss landscape, SAM [25] shows that dividing every gradient parameter with L2-norm to update will create a smooth path to possible optimum. Then, SAM restores the updated grad in the first step so the model can learn how to follow the same path to avoid harsh landscapes. Finally, the parameters were updated by the original gradient in the second step. With this operation, SAM has made a series of either residual networks or densely connective networks, such as EffNet-L2 [26] and PyramidNet [27], to gain the state-of-the-art performance in Imagenet, Cifar-10, and Cifar-100 classification benchmark.

Utilizing the attention mechanism in computer vision is also a norm. We have observed CBAM [28] used max pooling and average pooling to let convolutions capture different angles information to apply the pooling method. Among utilizing average pooling, SENets [7] used global average pooling to squeeze the spatial information into one value, and then it uses a simple multi-layer perceptron with a ReLU and another MLP layer with a Softmax to make channel attention. By embedding characteristic of SENet, the work showed an improvement toward embedding a SE-block after a convolutional layer in VGG [3], Inception Net [4, 5], and ResNeXt [29]. After, EfficientNet [26] proposed a general formula to help build a decent CNN-based structure and utilized similar SENets but with Swish [30] to obtain the state-of-the-art performance in that time. On the other hand, natural language processing has also seen a successful development with attention, especially the introduction of Transformer in [31]. Furthermore, ViT [18] arbitrary used the same multi-head attention in the Transformer to classify the Imagenet-1k picture. The same notion can also be seen in DeiT-B [32], which used attention to transfer the pre-trained parameter on image classification. In the work of BiT [8], we also can see that transferring parameters from a massive model has been another trend either in computer vision or natural language processing.

Except for EfficientNet and PyramidNet in finding a general convolutional structure formula, Wide ResNet [33] has revealed that expanding the width of a CNN layer can offer an efficient performance with increasing performance. Comparing different combinations of kernel size in two or three layers in a block, two layers give a robust performance in their experiments. Also, the order of stacking a batch normalisation, activation function, and convolution is a vital element in CNNs. PreAct ResNet [34] has proven to place batch normalisation and activation before the convolution can perform relatively well in most cases. Additionally, applying a bottleneck block is a popular method in big CNNs. Res2net [35] has proposed a different type of bottleneck to boost object detection performance. With the bottleneck structure in CNNs, the image model can reduce the parameters and maintain a deep structure. Sharing with the same notion, ShuffleNets [36] and Shufflenets v2 [37] used a channel shuffle operation after grouping convolutional layers to keep the same performance as the original CNNs.

**Critical analysis**
By ResNets and DenseNets, skip and densely-connection play significant roles in building deep structure in the field of computer vision. Attention mechanism has also been a trend. However, applying multi-head attention as ViT is inefficient to make attention global. The combination of kernels in CNNs is also a vital aspect. Learning from the Wide ResNets, wide CNNs can benefit more, so we designed a similar structure as the basic block in ResNets but in a wide version. Lastly, we are surprised by how efficiencies were ShuffleNets v1 and v2 used relative fewer parameters than ResNets, but they still maintained the performance as much as possible. Nonetheless, as the shuffle operation might mess up the memory location in the process of back-propagation, the saving time in computation was offset by grouping CNNs and re-building corresponding gradient direction.

## III. UPANets
In this section, the proposed methods are listed. The attention methods for channel-wise and pixel-wise are revealed, firstly. Then, the UPA block is shown after the attention. Combining the skip and densely-connection in the UPA block, an explanation of UPA layers shows how they work together in UPANets. The structure of UPANets is shown after the proposed methods of extreme connection.

**Channel pixel attention**
A convolutional kernel is good at capturing local information with learning weight in a kernel. Although a convolutional neural network can form a complex pattern by stacking deep enough layers, so it makes lower hidden layer process local information and deeper hidden layer capture global patterns, the process is not direct. Nonetheless, applying a network to learn the essential pixels from channel to channel in width might bring a positive effect and help CNNs consider global information directly. Therefore, we propose channel pixel attention, CPA, which applies a one-layer multi-layer perceptron (MLPs) to pay attention to the pixel in the same position across channels. The method can be presented as:

$$X = \sum_{c=1}^{n} x_c^R W_c^T + b \quad (1)$$

where $c$ indicates the channel[th], $X \in \mathbb{R}^{N \times C \times W \times R}$, $x_c^R \in \mathbb{R}^{N \times W \times H \times C}$, which is reshaped to do a dot product with $W_c^T$. $W_c^T \in \mathbb{R}^{N \times C \times C}$. After the pixel attention by one-layer MLP, batch normalization and layer normalization with residual connection are applied. The workflow of the CPA can be demonstrated in **Figure 2**. Moreover, the sample feature maps with demonstration are in **Figure 7**.

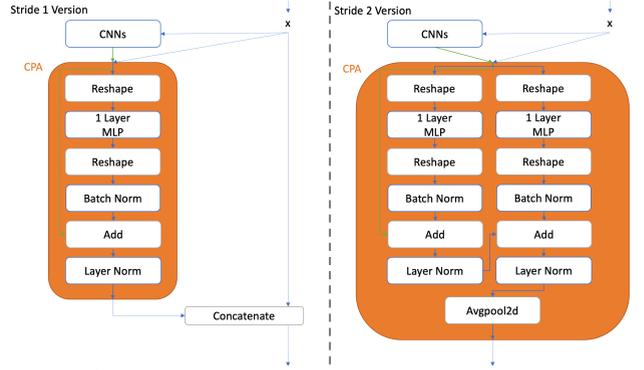

**Figure 2.** Channel pixel attention structure in stride one and stride two sets. In the orange region, CPA can make channel-wise pixel attention and downsample image by avgpool2d.

Among the CPA samples in **Figure 1**, the outputted feature maps from CPA are combining the original feature itself and helpful information from others. These combining feature show CPA can help a feature map fuse a more complex feature map without losing original features. Compared with deep structure, CPA helps a shallow network form complex pattern easily.

**Spatial pixel attention**
Global average pooling is widely applied in the image classification model. We agree that applied global average pooling before a final hidden layer can easily help the model learn which channel is vital for accuracy by weighing the representative value of a feature map. Most importantly, this operation does not require extra computational resource. However, we are wondering whether a learnable global pooling method could improve performance. To improve accuracy by important information in the spatial direction, we propose spatial pixel attention, SPA, which uses a one-layer perceptron. The method can be defined as the following formula:

$$X = \sum_{c=1}^{n} x_c^R W_c^T + b \quad (2)$$

where $c$ indicates the channel[th], $X \in \mathbb{R}^{N \times C \times 1}$, $x_c^R \in \mathbb{R}^{N \times C \times L}$, $L = W \times H$, and $W_c^T \in \mathbb{R}^{N \times L \times 1}$.

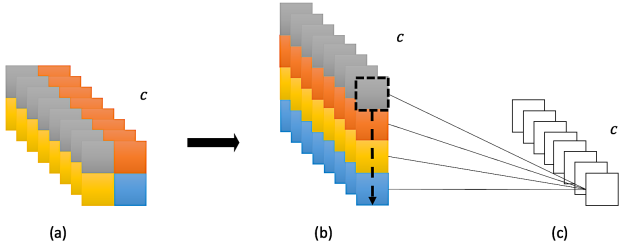

**Figure 3.** Spatial pixel attention. To demonstrate, we take a $2 \times 2$ feature map, in (a), with $c = 8$ as an example. Then, the process from (a) to (b) is reshaping the convolutional image. The (b) to (c) is applying spatial pixel attention, which is the same notion as the global average pooling.

In **Figure 3**, the process from (b) to (c) is implemented by a fully-connected neural network with a bias. By weighting a learnable matrix, SPA can decide to pay how much attention to essential pixels and then squeeze the whole pixel into one pixel by doing dot product instead of arbitrary pooling with average. In classifying Cifar-10 and Cifar-100, with $32 \times 32$ dimension per image, the maximum adding parameters is 1024 with no bias per feature map.

### Inverted triangular shape CNN layer with 3x3 kernels

Growing width in convolution is another helpful direction to improve performance. Also, the combination of two $3 \times 3$ convolutions is experimentally robust in most image classification. In UPANets, every first layer of CNN uses twice times channels of $3 \times 3$ kernel than the one. Thus, this shape can be viewed as an inverted triangle shape.

### UPA blocks

UPA blocks follow the findings in Wide ResNet which indicated the combination of two $3 \times 3$ convolutional layers could offer the most robust accuracy. The order of the convolution, batch normalization, and activation function follows the typical structure of CNNs. Meanwhile, CPA is applied parallelly, so the CPA input is the same as the CNN. Then, both outputs are simply added with layer-normalized afterwards. The structure can be seen in **Figure 4**.

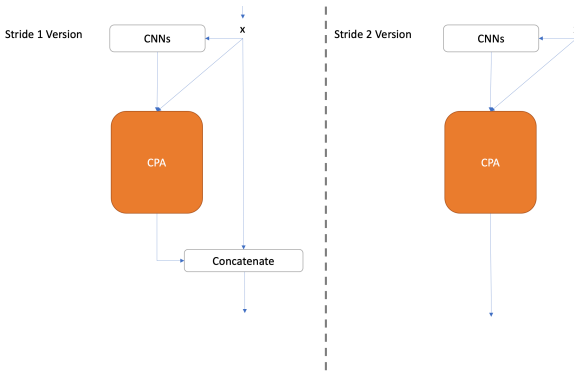

**Figure 4.** UPA blocks structure in the stride one and stride two sets.

From **Figure 4**, the differences between the stride one and stride two versions are applying to concatenate operation or not. The operation is densely connectivity. On the other hand, the residual connection is used in CPA to determine whether to output the current learned information or the information from the last block. Lastly, a $2 \times 2$ kernel average pooling is applied to down-sample; please referring **Figure 2.** By **Figure 4,** CPA can be embedded every CNNs-based models as SENets [7].

### UPA layers

In DenseNets, reusing features has been proved with a series of benefit, including reducing parameters, speeding up the computing process, and forming complex feature maps. This work uses densely-connection, but we modified it into a different UPA block structure, as **Figure 5.**

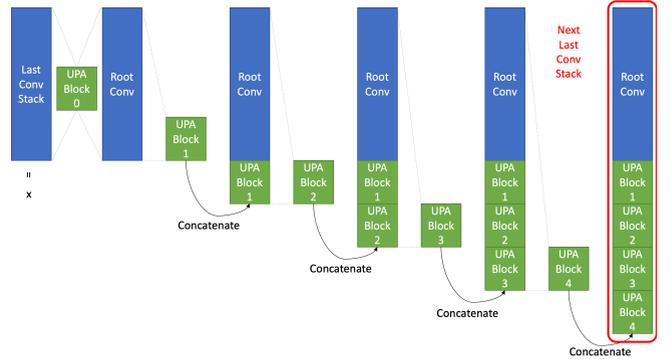

**Figure 5.** UPA layers with densely-connection. In the UPA block 0, a stride two UPA block uses the residual connection with $2 \times 2$ kernel average pooling is used.

The root information is preserved by the concatenating process until the last stride one UPA block. In the stride two UPA block, applying a $2 \times 2$ average pool means no stride two convolutions to down-sample. Except for the stride two operations in block 0 in every layer, each block follows the stride one operation. Nonetheless, the width of every stride one block is smaller than its input shape that can be referred to as the following equation:

$$w_b = W_l/b \qquad (3)$$

where $b = 1 \cdots n$, $W_l$ indicates the summation of adding width of this layer, $w_b$ indicates the output width of this block, and $w_0$ equals to two times width of the last layer because the original input is remained and the processed information is appended after that. For example, if the width of the layer 1 is set to 16, the outputted width of the layer 1 would be 32 because of densely-connection. Therefore, the block 0 width in the layer 2 is 32, $w_0 = 32$. Then, when the number of blocks in layer 2 is 4, $b = 4$, the width in every block is 8, $w_b = 8$ because $W_0 = 32$ and $\frac{32}{4} = 8$. In this case, the outputted width from this block of this layer will be 40.

**Table 1.** The UPANets structure for the Cifar-10. $N$ represents the data number, $F$ indicates the filters number, $B_i$ are blocks, $d$ means the depth multiplier, $b$ is the number of the block, and $w$ is the convolutional width. UPA Block 0 and the others Blocks follow the stride 2 and stride 1 UPA block, respectively.

| Layers | Blocks | Input size | Output size | Structure |
|---|---|---|---|---|
| UPA Root Layer | UPA Block 0 | $N \times 32 \times 32 \times 3$ | $N \times 32 \times 32 \times F$ | $\left\{ \begin{bmatrix} 3\times3\ conv, 2w \\ 3\times3\ conv, 1w \end{bmatrix} + CPA \right\}$ |
| UPA Layer 1 $B_1 = 4d$ | UPA Block 0 | $N \times 32 \times 32 \times F$ | $N \times 32 \times 32 \times \left(\left(\frac{F}{B_1}\right) + F\right)_{F_0}$ | $\left\{ \begin{bmatrix} 3\times3\ conv, 2w \\ 3\times3\ conv, 1w \end{bmatrix} + CPA \right\}$ |
| | UPA Block 1~4d | $N \times 32 \times 32 \times F_{b-1}$ | $N \times 32 \times 32 \times \left(\left(\frac{F}{B_1}\right) + F_{b-1}\right)_{F_b}$ | $\left\{ \begin{bmatrix} 3\times3\ conv, 2w \\ 3\times3\ conv, 1w \end{bmatrix} + CPA \right\} \times 4d$ |
| UPA Layer 2 $B_2 = 4d$ | UPA Block 0 | $N \times 32 \times 32 \times 2F$ | $N \times 16 \times 16 \times \left(\left(\frac{2F}{B_2}\right) + 2F\right)_{F_0}$ | $\left\{ \begin{bmatrix} 3\times3\ conv, 2w \\ 3\times3\ conv, 1w \end{bmatrix} + CPA, \\ Avgpool2d(stride\ 2) \right\}$ |
| | UPA Block 1~4d | $N \times 16 \times 16 \times F_{b-1}$ | $N \times 16 \times 16 \times \left(\left(\frac{2F}{B_2}\right) + F_{b-1}\right)_{F_b}$ | $\left\{ \begin{bmatrix} 3\times3\ conv, 2w \\ 3\times3\ conv, 1w \end{bmatrix} + CPA \right\} \times 4d$ |
| UPA Layer 3 $B_3 = 4d$ | UPA Block 0 | $N \times 16 \times 16 \times 4F$ | $N \times 8 \times 8 \times \left(\left(\frac{4F}{B_3}\right) + 4F\right)_{F_0}$ | $\left\{ \begin{bmatrix} 3\times3\ conv, 2w \\ 3\times3\ conv, 1w \end{bmatrix} + CPA, \\ Avgpool2d(stride\ 2) \right\}$ |
| | UPA Block 1~8d | $N \times 8 \times 8 \times F_{b-1}$ | $N \times 8 \times 8 \times \left(\left(\frac{4F}{B_3}\right) + F_{b-1}\right)_{F_b}$ | $\left\{ \begin{bmatrix} 3\times3\ conv, 2w \\ 3\times3\ conv, 1w \end{bmatrix} + CPA \right\} \times 4d$ |
| UPA Layer 4 $B_4 = 4d$ | UPA Block 0 | $N \times 8 \times 8 \times 8F$ | $N \times 4 \times 4 \times \left(\left(\frac{8F}{B_4}\right) + 8F\right)_{F_0}$ | $\left\{ \begin{bmatrix} 3\times3\ conv, 2w \\ 3\times3\ conv, 1w \end{bmatrix} + CPA, \\ Avgpool2d(stride\ 2) \right\}$ |
| | UPA Block 1~4d | $N \times 4 \times 4 \times F_{b-1}$ | $N \times 4 \times 4 \times \left(\left(\frac{8F}{B_4}\right) + F_{b-1}\right)_{F_b}$ | $\left\{ \begin{bmatrix} 3\times3\ conv, 2w \\ 3\times3\ conv, 1w \end{bmatrix} + CPA \right\} \times 4d$ |
| Ex-connected Layer | | $\begin{bmatrix} N\times32\times32\times F(Root\ Layer\ output), \\ N\times32\times32\times2F(Layer1\ output), \\ N\times16\times16\times4F(Layer2\ output), \\ N\times8\times8\times8F(Layer3\ output), \\ N\times4\times4\times16F(Layer4\ output), \end{bmatrix}$ | $N \times 1 \times 1 \times (F + F + 4F + 8F + 16F)$ | $SPA + GAP$ |
| Output Layer | | $N \times 1 \times 31F$ | $N \times 10$ | $Linear$ |

**Extreme connectivity**

Applying skip connection in a deep neural network has been a norm since ResNet introduced. Further, the dense connectivity in DenseNets has shown a different but more efficient way than before to connect the dense information. From the landscape of using skip connection, the surface is smoother, and thus this landscape raises the chance to reach a better optimum with a lower risk in overfitting. Based on this observation, to create an even smoother loss landscape, we introduce extreme connection; we will use exc in the following discussion across the whole model. It is only applied between each block and the last hidden layer. **Figure 6** eveals applied exc with SPA and global average pooling, GAP. This operation can be represented as the following:

$$X = F[SPA_1(x_1^R), SPA_2(x_2^R), \cdots, SPA_b(x_b^R)] \quad (4)$$

where $X \in \mathbb{R}^{N \times C}$, which is the output from the flatten-concatenate $F$. $N$ is the data number and $C$ represents the number of channels. Also, $b$ means the block[th] in a network. Different from the common image neural networks, which apply global average pooling before the final fully connected layer, we add the operation which combines SPA with GAP, as **Figure 6**:

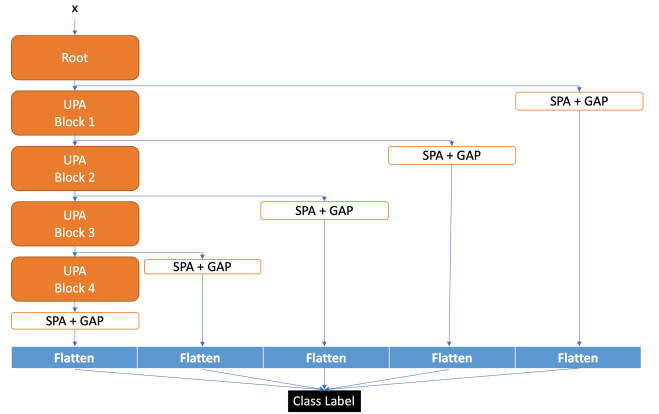

**Figure 6.** Extreme connection structure.

In **Figure 6**, exc builds the relationship from the final hidden layer to the output of each block. GAP servers the place of determining which convolution plays a vital role toward the label. SPA determines which pixel should be paid more attention to the class. By combining both operations with a layer normalization, both side information can be scaled to the same level to learn.

**UPANets structure**

**Table 1**, referring to the narrative in **UPA layers** the detail transferring of size, width, and the proposing attention in the Cifar-10, is presented. The proposed CPA is applied in each UPA block. Also, exc is used in every UPA layer with the proposed SPA and GPA.

## IV. EXPERIMENT

**Experiment environment and setting**

We implemented UPANets compared with CNN-based SOTAs for comparison. Although we do not reveal the costing time, it is better to unveil the experiment environment in a customer-based GPU, RTX Titan with 24GB, and an eight-core CPU, intel i9-9900KF, with 32GB RAM. As the limitation of the hardware, we mainly compared UPANets and others in Cifar-10, Cifar-100, and tiny Imagenet datasets. Every training process was implemented in a cosine annealing learning schedule with a half cycle. Similarly, every training optimizer was stochastic gradient descent with an initial learning rate of 0.1, momentum 0.9, and weight decay 0.0005. A simple combination of data argumentation was applied with random crop in padding 4, random horizontal flip, normalization, and input shape in Cifar and input shape in tiny Imagenet, respectively. As we conducted a series of experiments with different epochs, the specific used epochs number is revealed before in each sub-section experiment comparison. Lastly, the batch size was set to 100 in every training processes.

On the other hand, we used efficiency to examine the turnover rate between the parameters and accuracy throughout our experiments. Although the most crucial index is still the accuracy, also known as a top-1 error, we still hope the efficiency of the parameter should be considered during comparing models. The efficiency can be revealed as the following simple equation:

$$E = Acc/P \qquad (5)$$

where $E$ represents the efficiency, $P$ means the size of used parameters, and $Acc$ is the abbreviation of the accuracy. By this equation, we can learn whether this structure or setting could convert the parameters into performance efficiently. The meaning of the equation can also be understood as the ratio of accuracy and parameters. For example, if a 100% accuracy is brought by two parameters, $E = 0.5$. Also, if another 100% accuracy is contributed by four parameters, $E = 0.25$. By these two examples, the 0.5 is greater than 0.25 with the meaning of higher efficiency.

**Performance exploring in UPANets**

In this sub-section, we implemented a series of performan comparisons toward different components among UPANets. The performance of UPANets with $F = 16$ in Cifar-10 and Cifar-100 are revealed in the following comparisons, please see the meaning of $F$ in **Table 1**. Each performance was recorded in testing stage with the highest accuracy. The total epochs number in this sub-section was set to 100, and the experiment setting was also following the aforementioned experiment description in **Experiment environment and setting.**

**1. LEARNABLE EX-CONNECTION**

In the sub-section of **Extreme connectivity**, one of the reasons for ushering the connection is creating a smooth loss landscape to raise the chance to reach an optimum. Another reason is connecting a shallow layer with the final layer, and thus the model can be deep without facing overfitting. In the following table, we implemented UPANets16 in a series of variants. The variants were different in the connection structure. UPANets16 final GAP owns the typical CNN-based structure, which is only equipped with a GAP layer before the output layer. UPANets16 final SPA used SPA to replace the only GAP layer in typical CNNs. UPANets16 exc GAP follows the proposed exc structure with GAP layers. UPANets16 exc SPA shares the same structure as UPANets16 exc GAP but applied SPA layers instead. Lastly, UPANets16 (exc SPA & GAP) used layer normalizations to combine SPA and GAP layers with exc structure. The performance and efficiency of forenamed models are listed in **Table 2**.

**Table 2.** The performance comparison table among UPANets16 variants in Cifar-10 and Cifar-100.

| Model | Cifar-10 Acc % (Top 1 Error) | Cifar-100 Acc % (Top 1 Error) | Size (Million) (Cifar-10, Cifar-100) | Efficiency (Acc % / Million) (Cifar-10, Cifar-100) |
|---|---|---|---|---|
| UPA16 final GAP | 94.66 (0.0534) | 74.63 (0.2537) | (1.48, 1.53) | (63.95, 49.65) |
| UPA16 final SPA | 94.70 (0.0530) | 74.72 (0.2491) | (1.48, 1.50) | (63.98, 49.71) |
| UPA16 exc GAP | 94.75 (0.0525) | 74.60 (0.2540) | (1.48, 1.53) | (63.89, 48.83) |
| UPA16 exc SPA | 94.58 (0.0542) | 75.09 (0.2491) | (1.49, 1.53) | (63.68, 49.08) |
| UPA16 (exc SPA & GAP) | 94.9 (0.0510) | 75.15 (0.2485) | (1.48, 1.53) | (63.89, 49.11) |

**Table 2,** comparing the performances between UPANets16 final GAP and UPANets16 final SPA, shows that a learnable global average pooling by applying a fully-connected layer can improve the performance either in Cifar-10 and Cifar-100. The same trend is shown in the aspect of efficiency. However, when we ushered exc into UPANets16, UPANets16 exc GAP outperformed UPANets 16 exc SPA with better efficiency. As a result, we tried to apply layer normalization to combine both operations and then witnessed an improvement in Cifar-10 and Cifar-100. Also, efficiency became better. The evidence reveals that either GAP or SPA offers a specific contribution to improvement. The GAP can help to decide which combination of the channels is essential. Moreover, the combination of the pixels is essential among SPA. By combining both operations can supplement each other. The performance comparison toward whether using a fully-connected layer of

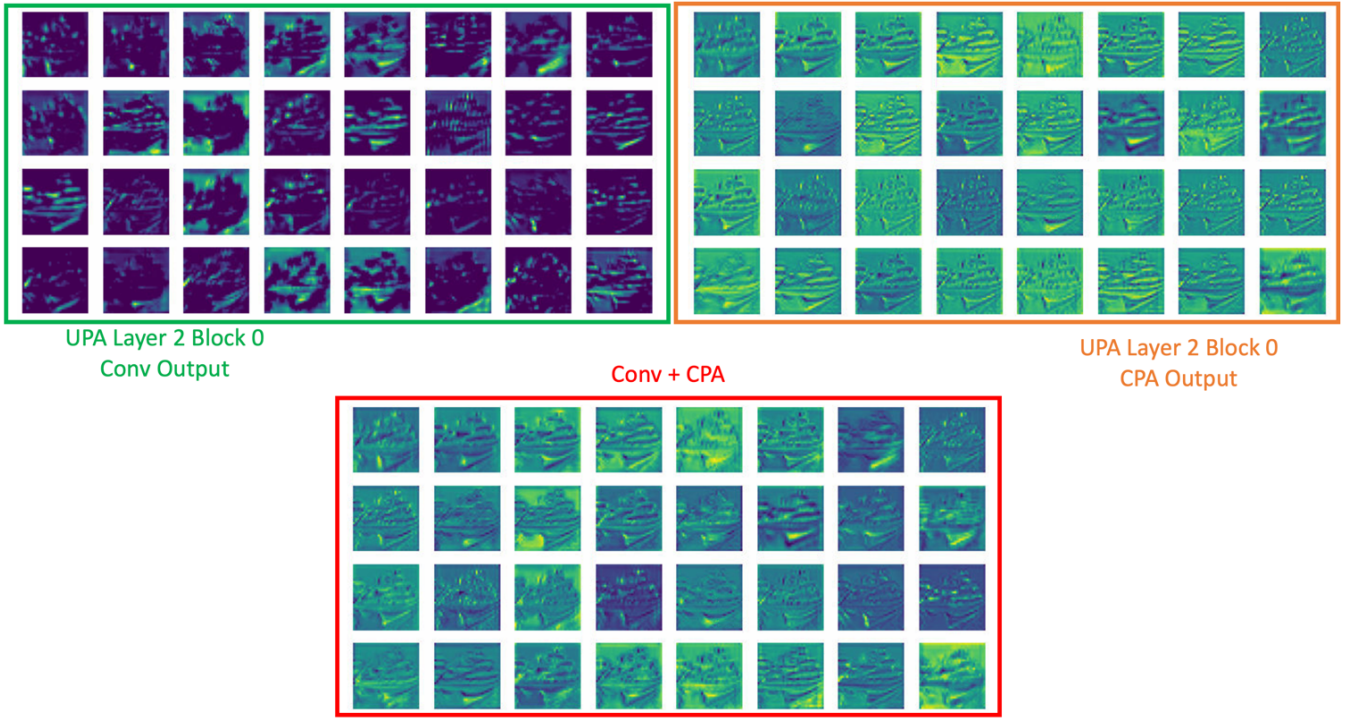

**Figure 7.** Samples of fusion feature maps in UPANets.

CNN layer in SPA can be seen in **A. CNN & Fully-connected layer comparison** in the Appendix.

We compared performance toward the accuracy, but we also followed the method in [16] with a slight modification to visualize different loss landscape in the same scale toward the loss of classifying Cifar-10. We used min-max scaling to convert different loss range into [0:1], which can be seen in **Figure 8**. Also, the top-1 error landscape is shown in **Figure 9**. As [16] explained, the landscape can only be regarded as the possible landscape for the visualization because it is produced by random sampling in a visual dimension. Regarding using min-max scaling for the loss landscape, an in-depth discussion is explained in **C. Landscape toward UPANets and Others** among **Appendix**.

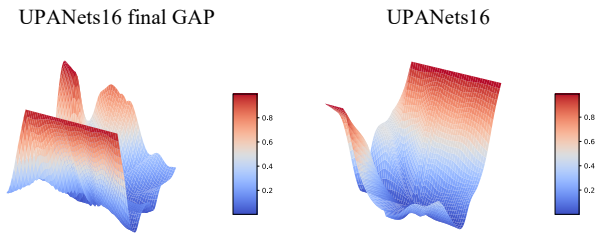

**Figure 8.** Normalizing loss landscape between UPANets16 final GAP and UPANets16.

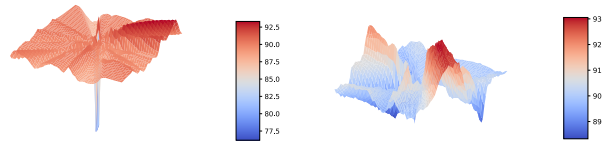

**Figure 9.** Top-1 error landscape toward UPANets16 final GAP and UPANets16.

The loss landscape in **Figure 8** and the top-1 error map in **Figure 9** illustrate that applying extreme connection did make the landscape smooth, so the chance of reaching minimum and preventing overfitting is rising. The difference between the original and normalized landscape becomes evident in the top-1 error landscape.

**2. FUSION OF CHANNEL PIXEL ATTENTION**
Based on the description of **Channel pixel attention**, we expect that this operation can help CNNs to consider global information as the widely used multi-head attention in the Transformer [31] but only needs one-third of parameters in attention by only using one fully-connected layer to do a weighted sum, instead of creating a query, key, and the value for attention. By **Figure 7**, we sampled the first 32nd feature maps from the convolution and the CPA layer of UPA Block 0 in UPA layer 2. The outputted feature maps are the information before using add and layer normalization, so the respective scale and output are remaining origin. We can see that the output of the CNN only detected a specific pattern toward the kernel. Also, some kernels only detected background information. Further, if the kernel could not detect a feature, a feature map remained dim. On the side of CPA outputs, every feature map covered the learned information from the others. Instead of simply extracting whole feature maps, each pixel considered the same position pixel from the others by learnable weights. Thus, the CPA can decide which pixel helps consider and vice versa. Before applying layer normalization, the samples of Conv + CPA own the detected pattern from the convolutional layer, local information, and concludes the global feature from other feature maps. The in-depth exploration of learned pattern in CNN and CPA can be seen in **D. Samples Pattern of the CNN and CPA in UPA block** of Appendix. In the bellowing **Table 3**, the improvement, which CPA brought, is discussed.

**Table 3.** The performance comparison table among UPA16 CPA variants toward Cifar-10 and Cifar-100.

| Model | Cifar-10 Acc % (Top 1 Error) | Cifar-100 Acc % (Top 1 Error) | Size (Million) (Cifar-10, Cifar-100) | Efficiency (Acc % / Million) (Cifar-10, Cifar-100) |
|---|---|---|---|---|
| UPA16 (w/o CPA) | 93.54 (0.0646) | 72.98 (0.2702) | (1.43, 1.48) | (65.32, 49.42) |
| UPA16 (groups=2, strides=2) | 94.56 (0.0544) | 74.99 (0.2501) | (1.29, 1.33) | (73.32, 56.20) |
| UPA16 shuffle (groups=2, strides=2) | 94.26 (0.0574) | 74.44 (0.2556) | (1.27, 1.31) | (74.37, 56.73) |
| UPA16 (groups=2) | 94.2 (0.058) | 74.52 (0.2548) | (1.02, 1.06) | (92.53, 70.12) |
| UPA16 shuffle (groups=2) | 93.33 (0.0667) | 71.98 (0.2802) | (0.96, 1.01) | (96.75, 71.31) |
| UPA16 (groups=4) | 93.79 (0.0621) | 73.55 (0.2645) | (0.78, 0.83) | (119.58, 88.73) |
| UPA16 shuffle (groups=4) | 92.93 (0.0707) | 71.33 (0.2837) | (0.73, 0.78) | (127.14, 91.97) |
| UPA16 | 94.90 (0.0510) | 75.15 (0.2485) | (1.48, 1.53) | (63.89, 49.11) |

In **Table 3,** UPANets16 w/o CPA reveals an obvious decease in both datasets so that CPA can boost the classification performance. On the other part, we also implemented a series of comparison among applying CPA and shuffle operation in ShuffleNets v1 and v2, as we realize CPA can offer the same effect of connecting independent CNNs. In that case, we want to validate whether the CPA can also maintain the same performance with fewer parameters. We placed the shuffle operation in the same place as ShuffleNets, which means there is a shuffle between two CNN layers with the first CNN in groups. In this experiment, CPA offered a better performance compared with shuffled UPANets. As the number of groups escalating, the performance difference between CPA and shuffle increases. While we agree that shuffle operation has very efficient parameters utilization, CPA can offer better performance with a minor resource trade-off.

**Results comparison with SOTAs**
UPANets was not only implemented in F=16, 32, and 64, a series of SOTAs were also reimplemented for comparison in Cifar-10 and Cifar100. The structure of reimplemented SOTAs followed the work in the link[1]. Every model was trained in 200 epochs and followed the experiment setting in **Experiment environment and setting**.

[1] https://github.com/kuangliu/pytorch-cifar

## 1. CIFAR-10

In this comparison, the performance of each model was recorded in accuracy toward testing data, parameters size in million and efficiency, as equation (5). Because there are three performance indexes in **Table 4**, we presented the information in a scatter plot as **Figure 10**, which contains accuracy in the y-axis and efficiency in the x-axis. The size of the circle toward each model represents a relative parameter size in a million compared with others. Besides, the specific used value for plotting and comparing can be seen in **Table 4**.

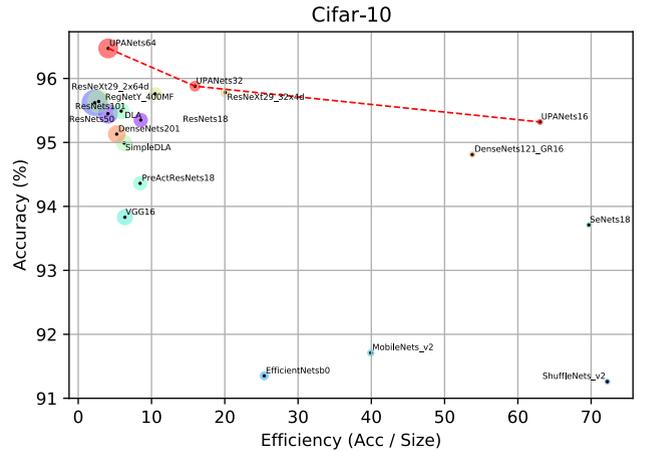

**Figure 10.** Scatter plot of UPANets performance with SOTAs in Cifar-10.

**Table 4.** The table of UPANets performance with SOTAs in Cifar-10.

| Model | Test Avg Accuracy ↑ | Size (M) | Efficiency |
|---|---|---|---|
| ShuffleNetsv2 | 91.26 | 1.26 | 72.21 |
| EfficientNetsb0 | 91.35 | 3.60 | 25.38 |
| MobileNetv2 | 91.71 | 2.30 | 39.93 |
| SeNets18 | 93.71 | 1.34 | 69.69 |
| VGG16 | 93.83 | 14.73 | 6.37 |
| PreActResNets18 | 94.36 | 11.17 | 8.45 |
| DenseNets121_16GR | 94.81 | 1.76 | 53.78 |
| SimpleDLA | 94.99 | 15.14 | 6.27 |
| DenseNets201 | 95.13 | 18.10 | 5.25 |
| UPANets16 | 95.32 | 1.49 | 63.03 |
| ResNets18 | 95.35 | 11.17 | 8.53 |
| ResNets50 | 95.45 | 23.52 | 4.06 |
| RegNetY_400MF | 95.46 | 5.71 | 16.71 |
| DLA | 95.49 | 16.29 | 5.86 |
| ResNets101 | 95.62 | 42.51 | 2.25 |
| UPANets32 | 95.88 | 5.93 | 15.93 |
| ResNeXt29_2x64d | 95.76 | 9.13 | 10.49 |
| ResNeXt29_32x4d | 95.78 | 4.77 | 20.06 |
| UPANets64 | 96.47 | 23.60 | 4.09 |

From **Figure 10** and **Table 4**, UPANets64 has the best accuracy. What is more, UPANets have an outstanding performance in balancing efficiency and accuracy in the scatter plot. We also observed that models claimed in the lite structure are located in the bottom right area, but they lost certain accuracy. Nonetheless, UPANets16 and DenseNets located in the upper right corner, indicating our proposed model and DenseNets have similar high efficiency. In terms of only viewing accuracy, UPANets64 is the only model reaching over 96% accuracy without needing too many parameters, especially compared with ResNets101 and DenseNets201.

2. CIFAR-100

We applied the same experimental setting with **1. CIFAR-10** in this Cifar-100 comparison. Similarly, please observe the result in **Figure** 11, corresponding with values in **Table 5**.

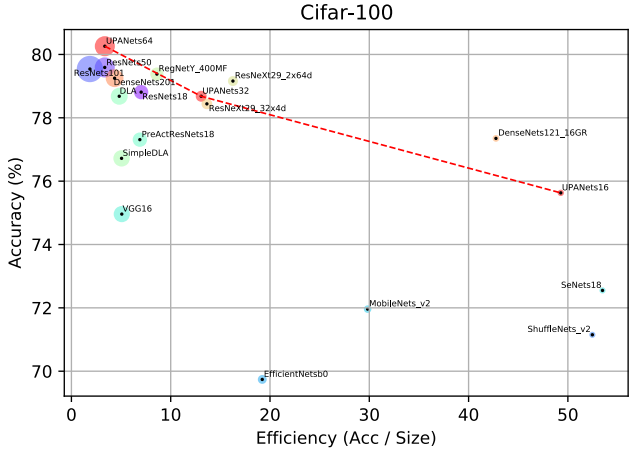

**Figure 11.** Scatter plot of UPANets performance with SOTAs in Cifar-100.

**Table 5.** Table of UPANets performance with SOTAs in Cifar-100.

| Model | Test Avg Accuracy ↑ | Size (M) | Efficiency |
|---|---|---|---|
| EfficientNetsb0 | 69.74 | 3.63 | 19.22 |
| ShuffleNetsv2 | 71.15 | 1.36 | 52.47 |
| MobileNetv2 | 71.96 | 2.41 | 29.83 |
| SeNets18 | 72.55 | 1.36 | 53.49 |
| VGG16 | 74.96 | 14.77 | 5.07 |
| SimpleDLA | 76.72 | 15.19 | 5.05 |
| UPANets16 | 76.73 | 1.56 | 49.05 |
| preactresnet18 | 77.31 | 11.22 | 6.89 |
| DenseNets121_16GR | 77.35 | 1.81 | 42.76 |
| RegNetY_400MF | 78.44 | 5.75 | 13.64 |
| DLA | 78.68 | 16.34 | 4.82 |
| UPANets32 | 78.78 | 6.02 | 12.90 |
| ResNets18 | 78.81 | 11.22 | 7.02 |
| ResNeXt29_32x4d | 79.16 | 4.87 | 16.27 |
| DenseNets201 | 79.25 | 18.23 | 4.35 |
| ResNeXt29_2x64d | 79.38 | 9.22 | 8.61 |
| ResNets101 | 79.54 | 42.70 | 1.86 |
| ResNets50 | 79.59 | 23.71 | 3.36 |
| UPANets64 | 80.29 | 23.84 | 3.37 |

By **Figure 11** and **Table 5**, UPANets64 also has the most excellent classification performance. Also, UPAnets variants had a decent performance as they surpassed most of SOTAs. The overall performance pattern is similar to **Figure 10**. So, we believe our UPANets has a competitive performance among classification tasks.

3. TINY IMAGENET

Although we compare a series of SOTAs with UPANets in Cifar-10 and Cifar100, the difficulty of datasets is relatively small comparing with Tiny Imagenet as it needs to classify two times more labels. Besides, the image size is also two times larger than Cifar-series datasets, so we only test UPANets64 in 100 epochs with the same experiment setting as comparison above. We compared with some SOTAs who also were tested on Tiny Imagenet in their works under below:

**Table 6.** Table of UPANets performance with SOTAs in Tiny Imagenet.

| Model | Test Avg Accuracy ↑ | Size (M) | Efficiency |
|---|---|---|---|
| DenseNets + Residual Networks [38] | 60.00 | N/A | N/A |
| PreActResNets18 [39] | 63.48 | N/A | N/A |
| UPANets64 | 67.67 | 24.40 | 2.77 |

Although it is still rare for comparing classification in Tiny Imagenet, we can know UPANets has not only excellent capability in simple datasets but also great ability in complex datasets like Tiny Imagenet. Our UPANets performance could be one of the state-of-the-art models in the Tiny Imagenet benchmark. Especially, a model which was trained end-to-end in a machine equipped with a customer-based GPU.

V. CONCLUSION

We proposed a new pixel-attention operation, CPA, which can capture global information and offer the same effect of Shuffle Nets with shallow depth and better accuracy. By ushering learnable global average pooling, SPA, and extreme connection, the smooth loss landscape can raise the chance of reaching minima. Integrating proposed methods into UPANets and comparing with a series of SOTAs in Cifar10, Cifar-100, and Tiny Imagenet, UPANets surpassed most SOTAs and can offer competitive performance in image classification. These evidence shows that learning universal pixels with proposed attention methods can profoundly improve computer vision ability.

## Appendix

### A. CNN & Fully-connected layer comparison

**Table 7**, UPANets16 (CNN) applied CNNs to replace all fully-connected layers in both CPA and SPA of UPANets16. While CNN can share weight in the spatial dimension, the benefit brought a side effect on performance. Similarly, the efficiency did not be dimmed by the extra parameters in the FC-UPANets16.

**Table 7.** The comparison of using CNN and Fully-connected layer.

| Model | Cifar-10 Acc % (Top 1 Error) | Cifar-100 Acc % (Top 1 Error) | Size (Million) (Cifar-10, Cifar-100) | Efficiency (Acc % / Million) (Cifar-10, Cifar-100) |
|---|---|---|---|---|
| UPA16 (CNN) | 94.76 (0.0442) | 74.85 (0.2515) | (1.48, 1.53) | (63.81, 48.93) |
| UPA16 (FC) | 94.90 (0.051) | 75.15 (0.2485) | (1.48, 1.53) | (63.89, 49.11) |

### B. Width in UPANets

From **Table 8**, the effect of width did bring positive performance, especially in a more difficult task as Cifar-100, though the efficiency decreased as the width going wider.

**Table 8.** The comparison of using different width CNNs in UPANets.

| Model | Cifar10 Acc % (Top 1 Error) | Cifar100 Acc % (Top 1 Error) | Size (Million) (Cifar10, Cifar100) | Efficiency (Acc % / Million) (Cifar10, Cifar100) |
|---|---|---|---|---|
| UPA16 (Slim) | 94.05 (0.0595) | 73.46 (0.2654) | (0.77, 0.82) | (121.47, 89.71) |
| UPA16 | 94.90 (0.051) | 75.15 (0.2485) | (1.48, 1.53) | (63.89, 49.11) |

### C. Landscape toward UPANets and Others

The introducing of the visualizing loss landscape method in [16] helps researchers understand the possible training landscape among the parameters of a model. By the description of the actual implementing source code[23], the primary usage is setting a random sampling range from -1 to 1 with a specific sampling number, and the default number is 50. However, using this strategy, as this sampling method is similar to the sensitivity analysis in determining feature importance, only proper sampling can produce a calculatable loss. This dilemma becomes even worse when we try to visualize a sensitive model, such as DenseNets, because a little adding noise might cause the loss to Nan. Therefore, how to define a good sampling range is a challenge. On the other hand, although the filter normalization has been introduced in [16] for comparing loss landscapes from different models, we found that different range of loss is still hardly comparing with others. An enormous total range of a loss will make most landscape smother because an outlier will break the harmony of the loss map. We used a grid search for finding a visualizable range carefully without modifying the original visualization method to address the previous barriers. On the ground of making two landscape comparable, we also used min-max scaling for every loss landscape. A series of before and after scaled landscapes are shown in the following figures. For demonstrating, we end-to-end trained a DenseNets and our models for Cifar-10 version based on the code in this project[4] and applied the method mentioned above in **Figure 12** and following comparisons.

---

[2] https://github.com/tomgoldstein/loss-landscape
[3] https://github.com/JoelNiklaus/loss_landscape

[4] https://github.com/kuangliu/pytorch-cifar

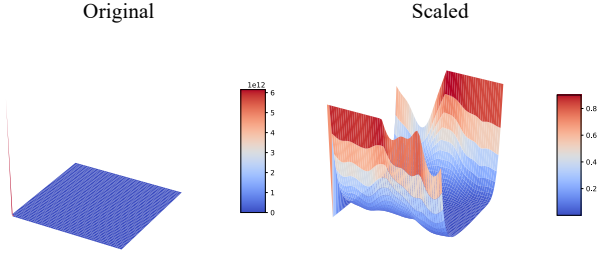

**Figure 12.** The loss landscape of un-scaled, left, and the scaled, right, of DenseNets.

What the visualizable sample range was $[-0.0375: 0.0375]$ with 50 samples. The largest loss broke the harmony of the original loss landscape on the left. The relative more minor loss owns the majority number, but it is hard to see the fluctuation of the landscape from the relative more minor loss because of the outlier. Therefore, we only see a flatten space on the left. Min-max scaled loss landscape shows a much different view on the right. Although the centre of the map is still flat, the surrounding loss stands erect on edge. Not only the scaled landscape can reveal a much reasonable profile, but scaling can also make different landscapes comparable. However, apart from the sampling range of DenseNets, the sample range among each UPANets variants was the same default range in [16], which is $[-1: 1]$. We offered UPANets16 loss and error landscapes, which are with and without scaled in the range $[-0.0375: 0.0375]$ in **Figure 13**. Please compare the original loss landscape in UPANets16 final GAP and UPANets16 in **Figure 8** and **Figure 14.**

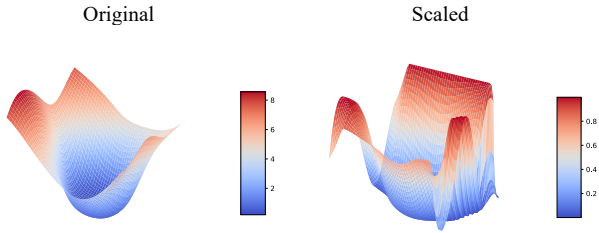

**Figure 13.** The UPANets16 loss landscape in the range $[-0.0375: 0.0375]$.

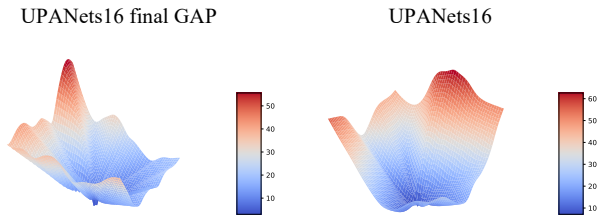

**Figure 14.** The original loss landscape of UPANets16 final GAP and UPANets16.

Apart from the loss landscape in UPANets16 final GAP and UPANets16, the loss landscape of the remaining models in **Table 2** are shown in the following figures. Then come with the figures of top-1 error landscape in **Figure 18**.

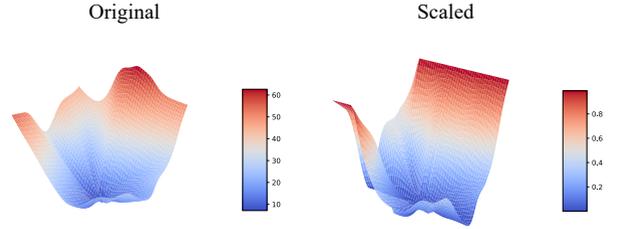

**Figure 15.** The original and scaled loss landscape of UPANets16 final SPA.

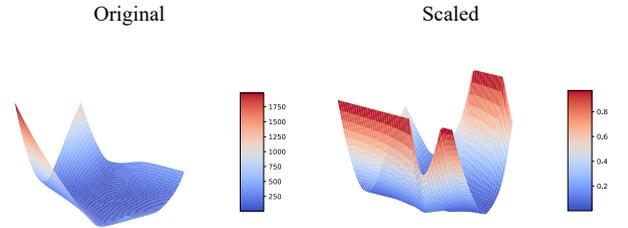

**Figure 16.** The original and scaled loss landscape of UPANets16 GAP.

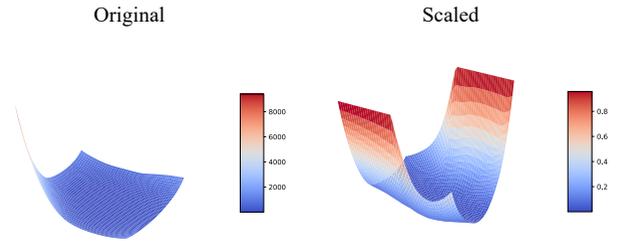

**Figure 17.** The original and scaled loss landscape of UPANets16 SPA.

By observing the scale bar on the right side of each plot, the ranges are different from landscape to landscape. Nonetheless, the min-max scaling makes every landscape comparable to the same level. From this series of scaled landscapes, we can further make sure that extreme connectivity offers a smother landscape compared with the landscapes of UPANets16 final GAP and SPA.

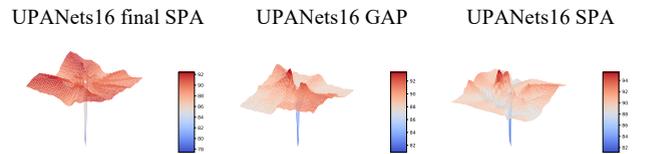

**Figure 18.** The top-1 error landscape of UPA 16 final SPA, UPA16 GAP, and UPA16 SPA.

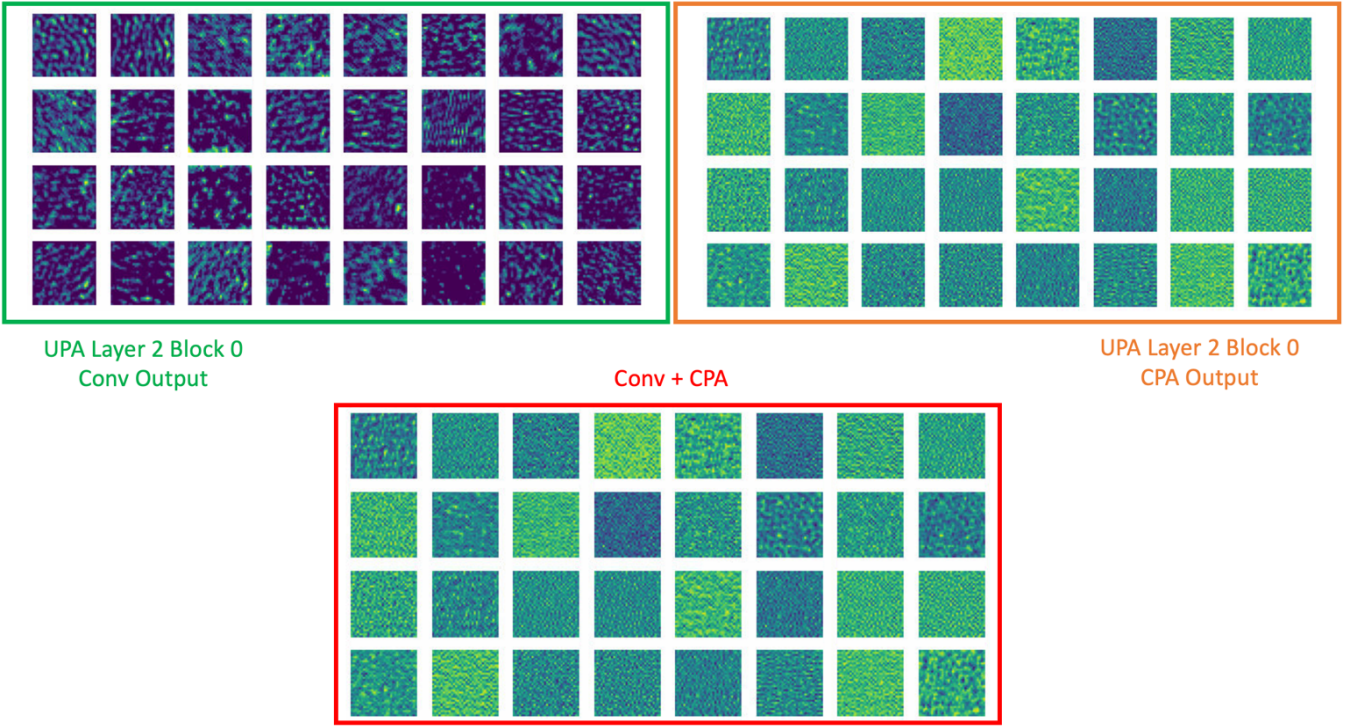

**Figure 19.** Samples of fusion feature maps in UPANets with using noise input.

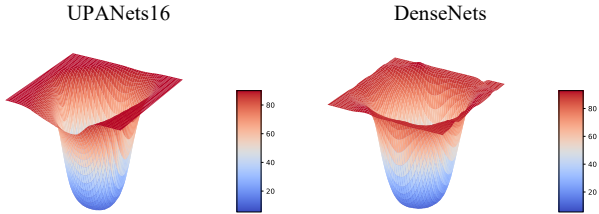

**Figure 20.** The top-1 error landscape of Cifar-10 version UPANets16 and DenseNets in the range $[-0.0375: 0.0375]$.

**Figure 18** and **Figure 20** show many different trends. No matter what version of UPANets16 variants in **Figure 18**, the top-1 error maps still present in a deep pattern. In contrast, the top-1 error map in UPANets16 and DenseNets show a smooth pattern, which is consistent with the observation in [16] and might be contributed by the dense connectivity. **Figure 20**, to compare in the same environment, contains the error landscape in the same range as **Figure 12** and **Figure 13**. We can observe that UPANets16 has the same smooth landscape as DenseNets.

### D. Samples Pattern of the CNN and CPA in UPA block

Following the same method in **2. fusion of channel pixel attention**, we sampled the feature maps with random noise, which follows the standard normal distribution. Thus, we can observe the actual convolution patterns and the forming complex CPA patterns in **Figure 19**. Without losing global information, the combination of convolution and CPA outputs also own detected local information.